%% file: main.tex
\documentclass[10pt,twocolumn,letterpaper]{article}

\usepackage{cvpr}              %
\usepackage{graphicx}
\usepackage{colortbl}
\usepackage{makecell}
\usepackage{enumitem}
\usepackage{balance}
\usepackage{bbding}
\usepackage{pifont}
\usepackage[T1]{fontenc}
\usepackage{calligra}
\usepackage{sidecap}
\usepackage{microtype}
\usepackage{adjustbox}
\usepackage{multirow}
\usepackage{array}
\usepackage{makecell}
\usepackage{amsmath}
\usepackage{amssymb}
\usepackage{booktabs}
\usepackage{float}
\usepackage{multirow}
\usepackage{mathtools}
\usepackage{amsthm}
\usepackage{caption}
\usepackage{widetext}
\usepackage{MnSymbol,wasysym}
\usepackage{CJKutf8}
\usepackage[utf8]{inputenc} %
\usepackage[T1]{fontenc}    %
\usepackage{amsfonts}       %
\usepackage{nicefrac}       %
\usepackage{microtype}      %
\usepackage{xcolor}         %
\usepackage{arydshln}

\usepackage[shortlabels,inline]{enumitem}

\usepackage{array}       %
\usepackage{inconsolata} %
\input{preamble}

\usepackage{amsfonts}
\definecolor{myAIBgColor}{RGB}{246,240,231}
\definecolor{myAIBgColorBack}{RGB}{140,131,118}
\input{mlVecMat}

\usepackage{alltt}
\tcbuselibrary{most}
\tcbset{
  aibox/.style={
    width=\textwidth,
    top=10pt,
    colback=myAIBgColor,
    colframe=myAIBgColorBack,
    colbacktitle=myAIBgColorBack,
    enhanced,
    center,
    attach boxed title to top left={yshift=-0.1in,xshift=0.15in},
    boxed title style={boxrule=0pt,colframe=white,},
  }
}
\newtcolorbox{AIbox}[2][]{aibox,title=#2,#1}
\definecolor{cvprblue}{rgb}{0.21,0.49,0.74}
\usepackage[pagebackref,breaklinks,colorlinks,allcolors=cvprblue]{hyperref}
\usepackage{subcaption} %
\usepackage[capitalize,noabbrev]{cleveref}
\crefname{figure}{Fig.}{Figs.}
\Crefname{figure}{Fig.}{Figs.}
\crefname{table}{Tab.}{Tabs.}
\Crefname{table}{Tab.}{Tabs.}
\crefname{section}{Sec.}{Secs.}
\Crefname{section}{Sec.}{Secs.}
\def\confName{CVPR}
\def\confYear{2026}
\usepackage{booktabs} 
\definecolor{lyellow}{HTML}{F7F2CC}
\definecolor{lpink}{HTML}{F6D6D6}
\definecolor{mycol1}{HTML}{F2EEEC}
\definecolor{mycol2}{HTML}{7E6A60}

\usepackage{algorithm}
\usepackage[noend]{algorithmic} %
\usepackage{amsmath} %

\renewcommand{\algorithmicrequire}{\textbf{Input:}}
\renewcommand{\algorithmicensure}{\textbf{Output:}}
\newcommand{\algorithmicparameter}{\textbf{Parameter:}}

\title{OmniZip: Audio-Guided Dynamic Token Compression for Fast Omnimodal Large Language Models}
\author{
    Keda Tao\textsuperscript{1,2,3,$^\dagger$},\quad Kele Shao\textsuperscript{1,4,2},\quad Bohan Yu\textsuperscript{3},\quad  Weiqiang Wang\textsuperscript{3},\quad Jian Liu\textsuperscript{3,$^*$},\quad Huan Wang\textsuperscript{2,$^*$} \\
    Zhejiang University\textsuperscript{1},
    Westlake University\textsuperscript{2},
    Ant Group\textsuperscript{3},
    Shanghai Innovation Institute\textsuperscript{4} \\
    \textcolor{magenta}{\tt{ \url{https://github.com/KD-TAO/OmniZip}}}
}

\begin{document}
\input{sec/teaser}

\input{sec/0_abstract}    
\input{sec/1_intro}

\input{sec/2_related}
\input{sec/3_method}

\input{sec/4_exp}

\section*{Acknowledgement}
This work was supported by the Ant Group Research Intern Program, Young Scientists Fund of the National Natural Science Foundation of China (NSFC) (No. 62506305), Zhejiang Leading Innovative and Entrepreneur Team Introduction Program (No. 2024R01007), Key Research and Development Program of Zhejiang Province (No. 2025C01026), Scientific Research Project of Westlake University (No. WU2025WF003), Chinese Association for Artificial Intelligence (CAAI) \& Ant Group Research Fund - AGI Track (No. 2025CAAI-ANT-13). It is also supported by the research funds of the National Talent Program and Hangzhou Municipal Talent Program.

{
    \small
    \bibliographystyle{ieeenat_fullname}
    \bibliography{main}
}

\input{sec/X_suppl}

\end{document}

%% file: preamble.tex
\newcommand{\red}[1]{{\color{red}#1}}

%% file: mlVecMat.tex
\usepackage{amsmath}
\usepackage{amsfonts}
\usepackage{amssymb}
\usepackage{wrapfig}
\usepackage{subcaption}
\usepackage{multirow}
 \usepackage{mathtools} 

\usepackage{verbatim}

\usepackage{anyfontsize}

\usepackage{microtype}
\usepackage{graphicx}
\usepackage{booktabs} %
\usepackage{multirow}
\usepackage{amsmath,amssymb}
\usepackage{booktabs}
\usepackage{caption,subcaption}

\usepackage{xcolor}
\definecolor{mygreen}{HTML}{3cb44b}
\definecolor{skyblue}{HTML}{beffff}
\definecolor{lightgreen}{HTML}{90ee90}

\usepackage{color, colortbl}

\definecolor{emerald}{rgb}{0.31, 0.78, 0.37}

\usepackage{tcolorbox}
\usepackage{enumitem}
\setitemize{itemsep=10pt,topsep=0pt,parsep=0pt,partopsep=0pt}
\usepackage{colortbl}

\usepackage{xcolor}
\definecolor{mygreen}{HTML}{3cb44b}
\colorlet{myyellow}{green!10!orange!90!}
\makeatletter

\usepackage{tikz}
\usetikzlibrary{arrows,shapes,snakes,automata,backgrounds,fit,petri}
\usepackage{adjustbox}

\newcommand{\RN}[1]{%
	\textup{\lowercase\expandafter{\it \romannumeral#1}}%
}
\usepackage{tabu}

\newcommand{\beq}{\vspace{0mm}\begin{equation}}
\newcommand{\eeq}{\vspace{0mm}\end{equation}}
\newcommand{\beqs}{\vspace{0mm}\begin{eqnarray}}
\newcommand{\eeqs}{\vspace{0mm}\end{eqnarray}}
\newcommand{\barr}{\begin{array}}
\newcommand{\earr}{\end{array}}

\usepackage{color, colortbl}
\definecolor{Gray}{gray}{0.93}

\usepackage{lipsum}

\usepackage{pifont}%

\usepackage{makecell}

\usepackage{xcolor,amsmath}
\usepackage[linesnumbered,ruled,vlined]{algorithm2e}
\DontPrintSemicolon

\usepackage{xcolor}
\definecolor{mygreen}{HTML}{3cb44b}

\SetKwComment{Comment}{\color{green!50!black}\# }{}

\SetKwProg{Function}{def}{:}{}

\SetKwProg{For}{for}{:}{}
\SetKwProg{If}{if}{:}{}

%% file: sec/teaser.tex
\twocolumn[{
\renewcommand\twocolumn[1][]{#1}
\maketitle
\begin{center}
\centering
\vspace{-2mm}
\captionsetup{font={small}, skip=3pt}
\includegraphics[width=1\linewidth]{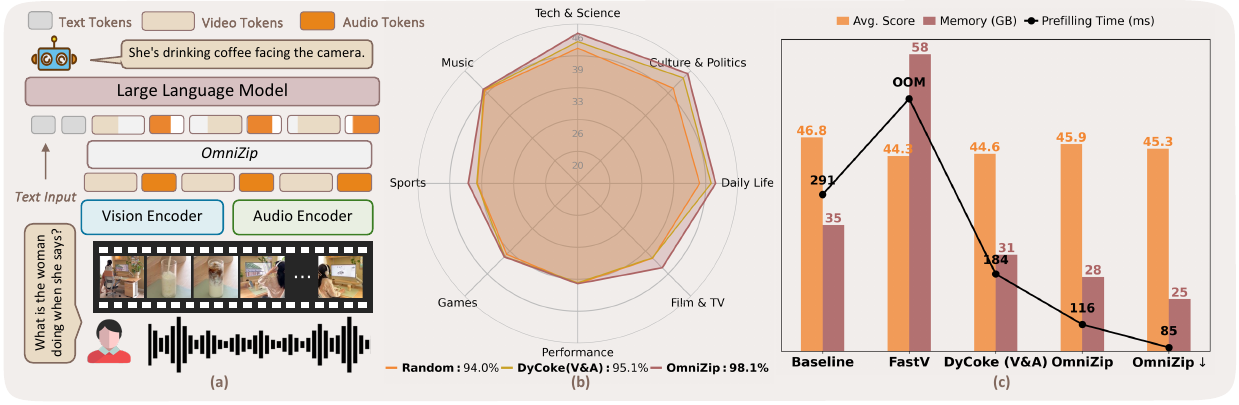}
\captionof{figure}{\textbf{(a)}: We introduce \textit{OmniZip}, an audio-video token compression method tailored for efficient OmniLLMs. The key innovation is a ``\textit{listen-to-prune}'' paradigm -- utilizing \textit{audio} to dynamically guide video token pruning, complemented by a proposed compression module. \textbf{(b)}: OmniZip achieves superior performance on various audio-video tasks on WorldSense~\cite{hong2025worldsense}, outperforming other methods. \textbf{(c)}: Efficiency and performance comparison on WorldSense with Qwen2.5-Omni~\cite{xu2025qwen2}. OmniZip can achieve 2.51-3.42$\times$ wall-clock inference speedup (on an A6000 48G GPU), 1.4$\times$ memory reduction against other top-performing methods with almost the same performance.
}
\label{fig:teaser}
\end{center}
}]

%% file: sec/0_abstract.tex
\begin{abstract}
Omnimodal large language models (OmniLLMs) have attracted increasing research attention of late towards unified audio-video understanding.
However, the high computational cost of processing longer joint audio-video token sequences has become a key bottleneck. 
Existing token compression methods have not addressed the emerging need to jointly compress multimodal tokens.
To bridge this gap, we present \textbf{OmniZip}, a training-free, audio-guided audio-visual token-compression framework that optimizes multimodal token representation and accelerates model inference.
Specifically, OmniZip first identifies salient audio tokens, then computes an audio retention score for each time group to capture information density, thereby dynamically guiding video token pruning and preserving cues from audio anchors enhanced by cross-modal similarity. 
For each time window, OmniZip compresses the video tokens using an interleaved spatio-temporal scheme.
Extensive results demonstrate the merits of OmniZip: it achieves a 3.42$\times$ inference speedup and a 1.4$\times$ memory reduction over other top-performing counterparts, while maintaining the performance of OmniLLMs without training.

\renewcommand{\thefootnote}{\fnsymbol{footnote}}
\makeatletter
\renewcommand{\@makefntext}[1]{%
  \parindent 1em\noindent
  \hbox to 1.8em{#1\hss}} %
\makeatother
\footnotetext[1]{$^*$Corresponding authors: Huan Wang (\texttt{wanghuan@westlake.edu.cn}), }
\footnotetext[2]{Jian Liu(\texttt{rex.lj@antgroup.com}).}
\footnotetext[3]{$^\dagger$Work done during internship at Ant Group.}

\end{abstract}

%% file: sec/1_intro.tex
\section{Introduction}
\label{sec:intro}

Video large language models (VideoLLMs) have demonstrated strong performance in video question answering and complex scene understanding~\cite{li2023videochat,lin2023video,liu2023visual,zhang2023video,bai2025qwen2,zhang2025videollama,wang2024qwen2,chen2024internvl,team2025gemma,llava-ov}. 
Due to a video inherently containing both visual and auditory streams, recent efforts have begun to focus on \textit{omnimodal large language models} (OmniLLMs) towards unified audio–video understanding~\cite{tang2025video,zhang2024videoin,xu2025qwen2,xu2025qwen3,yang2025humanomniv2,ge2025arc,shu2025audio,ye2025omnivinci,tong2025interactiveomni,li2024baichuanomni,fu2024vita,li2025omnivideobench}.

However, OmniLLM inference at scale remains constrained by the computational and memory bottleneck, primarily due to the prohibitively large number of audio-video tokens and the quadratic complexity of attention in large language models \cite{tao2025dycoke,shao2025holitom,shao2025tokens,shen2025fastvid}. 
Token compression techniques have been a promising approach for facilitating long-sequence inference on multimodal LLMs.
Recent works have been investigating token reduction from a \textit{purely visual} perspective~\cite{chen2024image,xing2024pyramiddrop,ye2025fit,yang2025topv,tan2025tokencarve,shen2024longvu,tao2025dycoke,shao2025holitom,huang2024prunevid,shang2024llava,yang2025visionzip,shen2025fastvid,shao2025tokens,chen2025streamingtom,liu2025revisiting}, while for OmniLLMs, the additional \textit{audio} tokens further inflate sequence length and are non-negligible.
At the core of technical challenges, audio and video streams exhibit distinct temporal scales and varying sparsity, and the coexistence of redundancy and complementarity renders token pruning particularly sensitive and challenging.
As such, \emph{joint audio–video token compression for OmniLLMs remains underexplored so far.}

\begin{figure}[t]
\centering
\captionsetup{font={small}, skip=-1pt}
\includegraphics[width=1\linewidth]{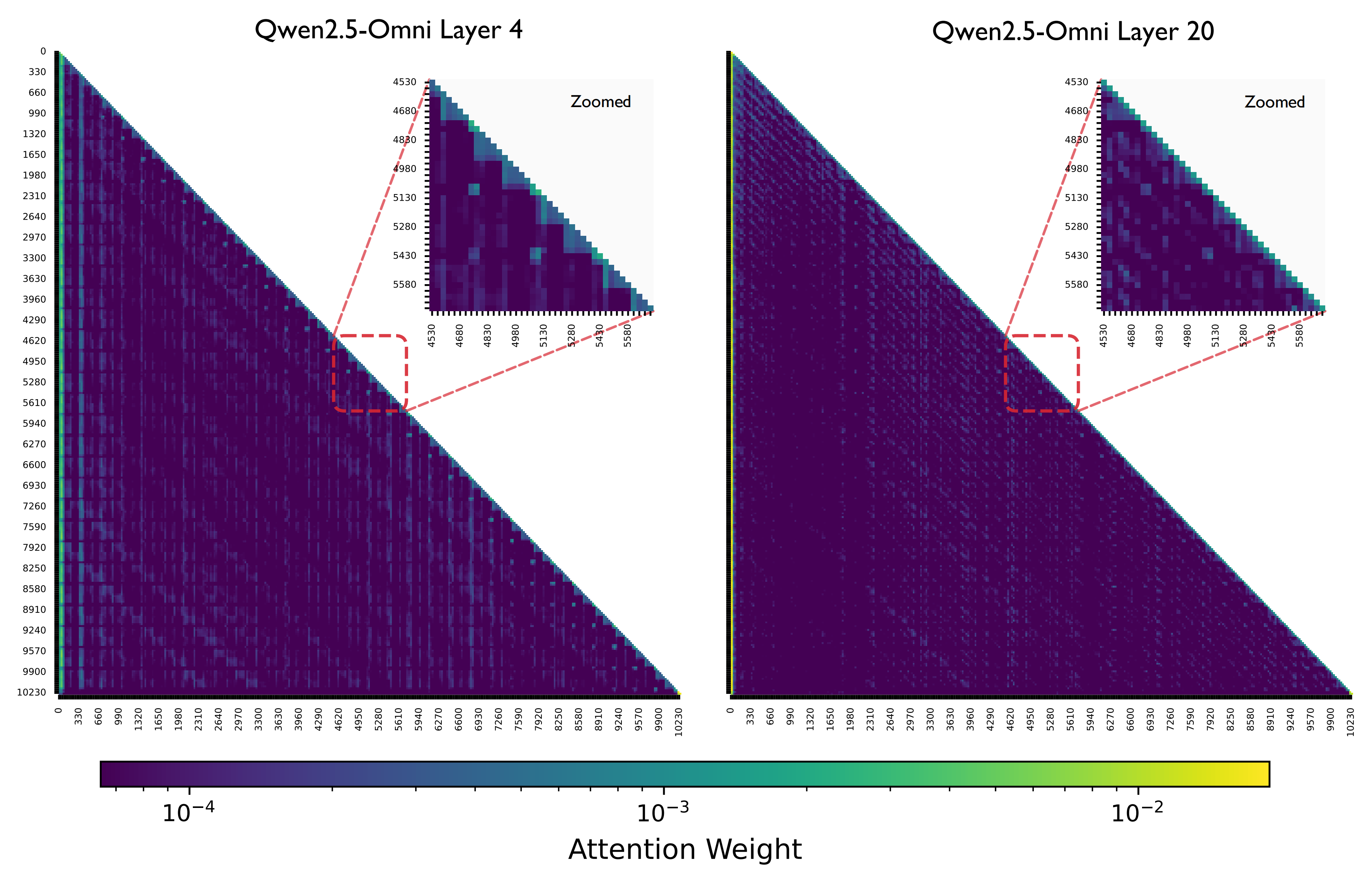}
\caption{\textbf{Audio tokens dominate attention heatmaps.} Regular vertical bands aligned with audio-token positions indicate consistently higher attention to audio tokens, while many video tokens receive little attention, suggesting greater redundancy. Attention aggregates within time windows and decays across windows, indicating that audio and video tokens preferentially attend to short-range context within the same window. Moreover, deeper layers allocate less attention to raw audio and video tokens.}
\label{fig:attn_v}
\vspace{-5mm}
\end{figure}

This work presents, to our knowledge, the first systematic study of reducing tokens under omnimodal inputs, and we propose \emph{OmniZip}, an audio-guided audio-video token compression method for OmniLLMs, as shown in \cref{fig:teaser}(a). 
Specifically, we start by performing token attention analyses.
In OmniLLM, the token sequence is constructed by segmenting the audio and video streams into fixed-length time windows. 
\cref{fig:attn_v} shows regularly recurring vertical bands at audio-token positions, indicating that attention on audio tokens is consistently greater than on video tokens, which suggests the dominance of audio inputs. 
A magnified view indicates predominantly intra-window attention—mutual attention between audio and video tokens within the same window is most pronounced. 
This pattern suggests that token compression should operate at the time-window granularity, which differentiates from prior single-modal compression strategies. 
A detailed analysis appears in \cref{sec:3_2_ana}.

Based on these analyses, OmniZip features three novel technical innovations. 
\textit{First}, we identify dominant audio tokens and compute the audio retention rate for each time window, which we interpret as time-wise information density and an event-boundary prior.
\textit{Second}, windows with high retention are treated as information-dense, and the corresponding video tokens receive a lower pruning rate; conversely, information-sparse windows are assigned higher pruning rates. 
\textit{Third}, to further preserve multimodal capability, we uniformly sample audio anchors and select secondary audio tokens for merging via cross-modal similarity. 
For video tokens, we propose an interleaved spatio-temporal token compression method that addresses temporal redundancy between frames and spatial redundancy within frames.
This interleaved design suppresses redundancy while avoiding excessive reduction along any single dimension.

Empirically, OmniZip demonstrates strong performance on audio-video understanding tasks, significantly outperforming single-modality token compression methods.
As shown in \cref{fig:teaser} (c), OmniZip achieves a 2.51$\times$ to 3.42$\times$ inference speedup on Qwen2.5-Omni-7B \cite{xu2025qwen2}, all while exhibiting the lowest memory consumption (reducing the GPU memory footprint by 10G) and maintaining the highest accuracy.
Crucially, OmniZip is \textit{training-free}.

Our contributions in this work are summarized as follows:
\begin{itemize}[itemsep=2pt]
    \item This work presents, to our knowledge, the first analysis of how audio-video tokens can be pruned to reduce computational overhead in omnimodal settings, and proposes OmniZip, a novel, training-free audio-video token compression framework for OmniLLMs to accelerate inference.
    \item We propose an audio-guided token compression method, complemented by a proposed video token compression module, to aggressively prune audio-video tokens while preserving cross-modal semantic and temporal alignment.
    \item Experimental results on several audio-video understanding benchmarks show that OmniZip can compress audio-video tokens while maintaining high inference accuracy, significantly improving inference speed, and reducing memory overhead.
\end{itemize}

%% file: sec/2_related.tex
\section{Related Work}
\label{sec:rel}

\subsection{Omnimodal Large Language Models}
To achieve a more human-like multimodal interaction experience, OmniLLMs have emerged. By leveraging multimodal data, they learn richer contextual information and achieve a deeper understanding of inter-modal relationships~\cite{xu2025qwen2,tong2025interactiveomni,xu2025qwen3,xie2024mini,tang2025video,zhang2024videoin,yang2025humanomniv2,ge2025arc,fu2024vita,shu2025audio,sun2024video,li2024baichuanomni}. 
In video understanding tasks, compared to VideoLLMs, OmniLLMs can additionally consider audio information alongside visual data, enabling more realistic answers and a more comprehensive understanding. 
Recent work, such as Qwen2.5-Omni~\cite{xu2025qwen2}, introduced an end-to-end model capable of perceiving all modalities. 
While InteractiveOmni~\cite{tong2025interactiveomni} has enabled multi-round audio-video conversations, significant recent work~\cite{ai2025ming,xu2025qwen3,ye2025omnivinci,yang2025humanomniv2} has further advanced state-of-the-art omnimodal understanding capabilities.
However, the large number of multimodal tokens introduced by video and audio inputs significantly impedes the practical deployment and application of OmniLLMs. 
Balancing model performance and computational efficiency remains a significant challenge. 
Thus, developing efficient methods to simplify the derivation of tokenized audio-video information is essential.

\subsection{Token Compression}
Recent research has focused on token compression to enhance the inference efficiency of multimodal large language models. 
This approach is highly effective as multimodal inputs often contain significant redundancies, such as image~\cite{bolya2022token,shang2024llava,yang2025visionzip,chen2024image,xing2024pyramiddrop,ye2025fit,yang2025topv,tan2025tokencarve}, video~\cite{chen2025streamingtom,tao2025dycoke,shao2025holitom,huang2024prunevid,shen2025fastvid,shen2024longvu}, and audio~\cite{li2023accelerating,lin2024speechprune,lee2025token,sun2024video}. 
A key advantage is that these methods can be applied as a tuning-free, post-processing technique. 
These methods operate by first establishing a metric to evaluate token importance, followed by corresponding compression operations~\cite{shao2025tokens}. 
While token compression methods for single modalities have been widely studied, their application to the omnimodal setting has not yet been explored. 
Furthermore, current mainstream methods typically depend on accessing the attention matrices from either the video encoder or the LLM~\cite{tao2025dycoke,shao2025holitom,yang2025visionzip,he2024zipvl,xing2024pyramiddrop}. 
This dependency is often incompatible with modern optimizations such as FlashAttention~\cite{dao2022flashattention,dao2023flashattention2}, necessitating the materialization of the full attention matrix. 
In conjunction with ultra-long visual token sequences, this readily leads to Out-of-Memory (OOM) errors. Therefore, such methods exhibit poor scalability to larger, more advanced models.    
Considering the inherent coupling of video and audio, we conduct the first exploration of token compression for the combined audio-video understanding task, aiming to facilitate the practical deployment of OmniLLMs.

%% file: sec/3_method.tex
\section{Proposed Method}
\label{sec:med}
\begin{figure*}
\centering
\captionsetup{font={small}, skip=0pt}
\includegraphics[width=1\linewidth]{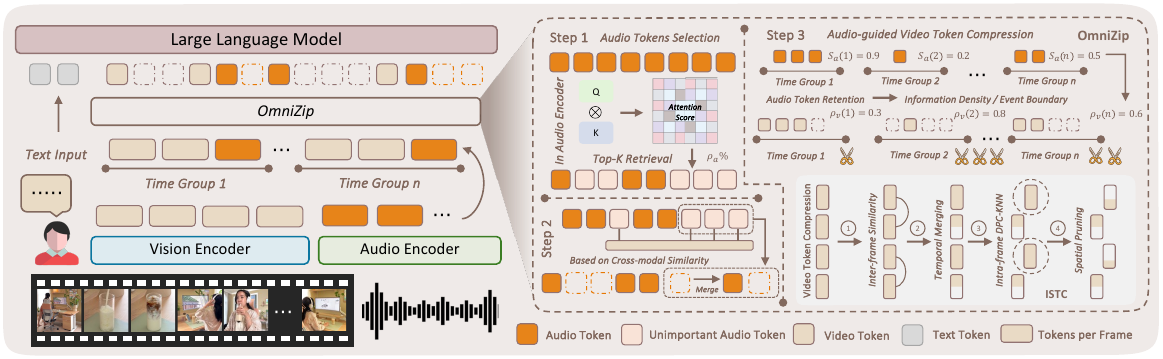}
\vspace{-2mm}
\caption{\textbf{Detailed overview of our OmniZip method.} 
First, OmniZip computes an audio retention rate derived from dominant audio tokens to determine a dynamic pruning rate for the corresponding video tokens. Next, to preserve multimodal information, we uniformly sample audio anchors and merge with non-anchor tokens selected via cross-modal similarity. Finally, video tokens undergo interleaved spatio-temporal compression (ISTC), which alternately reduces temporal redundancy by merging cross-frame tokens and spatial redundancy by pruning intra-frame tokens. $\rho_a$ is the compression ratio of the audio token, $S_a(i)$ and $\rho_v(i)$ are the audio token retention ratio and video token compression ratio, in each time group, respectively.} 
\label{fig-MED}
\vspace{-4mm}
\end{figure*}

In this section, we first describe the overall architecture of OmniLLMs (\cref{sec:2.1}), and then present the analyses based on the token attention distributions (\cref{sec:3_2_ana}). 
Next, we detail our proposed method, OmniZip (\cref{sec:2_omnizip}).
Then, the ISTC module for video-token pruning is introduced. \cref{fig-MED} illustrates the overall architecture.
Finally, we further remark on the design concept of our method in \cref{sec:3_remark}.

\subsection{Background on OmniLLM}
\label{sec:2.1}
OmniLLMs aim to ingest a full range of modalities together with human-provided prompts to form a unified audio–video understanding. 
Such models typically comprise a vision encoder, an audio encoder, a projector, and an LLM backbone.
Given a video, we first decompose it into individual video frames clip $X_{\text{vid}}\in\mathbb{R}^{T\times H\times W\times3}$ and audio segments $X_{\text{aud}}$ sampled at fixed rates, where T is the number of frames after sampling.
The vision encoder $g_{v}$ and audio encoder $g_a$ convert the raw video clip and audio clip into a sequence of token embeddings:
\begin{equation}
        \mathbf{Z}_{\text{v}} = g_{v}\!\left(X_{\text{vid}}\right),
        \quad
        \mathbf{Z}_{\text{a}} = g_{a}\!\left(X_{\text{aud}}\right),
\end{equation}
where $\mathbf{Z}_{\text{v}}\in\mathbb{R}^{N_v \times D}$, $\mathbf{Z}_{\text{a}}\in\mathbb{R}^{N_a \times D}$, $N_a$ and $N_v$ are the number of audio tokens and video tokens, respectively.
Then, the projector maps audio-video tokens into the LLM’s embedding space, enabling the model to process multimodal inputs effectively.
Typically, a video yields 10–20k tokens (audio and video), severely constraining efficient deployment.

Furthermore, the stitching for audio–video tokens is organized by fixed-length time windows, as shown in \cref{fig-MED}. 
The audio and video streams are segmented into multiple windows of equal duration. 
Within each window, co-temporal multimodal tokens are aligned and concatenated into a cross-modal block; the blocks are then concatenated chronologically to form a long token sequence and fed to the LLM. 
The LLM jointly aligns video, audio, and textual representations to generate a response.

\subsection{Token Attention Analysis}
\label{sec:3_2_ana}
To characterize redundancy and attention patterns in audio and video tokens during inference, we visualize the attention distribution, as shown in \cref{fig:attn_v}. 
First, most tokens receive a low attention score, and the attention to both video and audio tokens decreases with layer depth, indicating that judicious token pruning can preserve model reasoning while reducing memory usage and accelerating inference.

Then, we investigate how to design an effective token compression strategy. 
First, we observe regularly recurring bright bands in the attention heatmap. 
Cross-referencing with token indices shows that these bands align with audio tokens in each time window. 
This indicates that audio tokens are consistently assigned greater attention than video tokens across layers, whereas large regions of video tokens exhibit significantly lower attention scores, suggesting substantial redundancy and the dominant role of audio tokens in the inference process.
Magnified views reveal block-structured local attention: tokens cluster strongly within the same time window but decay rapidly across windows, indicating a strong locality for short-range temporal dependence.
This motivates us to design OmniZip to perform token pruning separately within each time window. 

Building on these observations, we design an audio-guided dynamic compression strategy for audio-video tokens.
Specifically, after selecting retained audio tokens, we treat per-window audio retention as a proxy for information density and event-boundary likelihood, and we dynamically allocate the video pruning rate for each time window accordingly, while constraining the video compression to exceed the audio compression. 
This reduces the number of tokens processed downstream while preserving performance, substantially lowering computational and memory costs.

\subsection{Our Method: OmniZip}
\label{sec:2_omnizip}
OmniZip is a \emph{training-free}, inference-time compressor that selects and restructures audio–video tokens before feeding them to the LLM.
As shown in \cref{fig-MED}, it proceeds window-by-window and contains three stages: (i) audio token selection, (ii) audio anchor consolidation, and (iii) audio–guided dynamic video compression.
Let the $t$-th time window contain $n_a$ audio tokens and $n_v$ video tokens, with embeddings $H_{a}^{t,i},H_{v}^{t,j}$ after the projectors.

\vspace{0.2em}
\noindent \textbf{Audio Token Selection.}
We filter audio tokens based on the attention distribution produced by the audio encoder. Specifically, we use the last layer of the audio encoder $g_a$ and compute the attention matrix:
\begin{equation}
A=\mathrm{Softmax}(QK^T/\sqrt{d})\in\mathbb{R}^{B\times N_a\times N_a},
\end{equation}
where $\mathbf{Q},\mathbf{K}\in\mathbb{R}^{N_a\times d}$ are the query and key matrices for the audio tokens, and $d$ is the state dimension. We quantify token importance as the mean attention each audio token receives from all other audio tokens, yielding a per-token score vector $a_{avg}\in R^{B\times N_a}$.
Tokens with larger mean-attention scores are considered more salient. 
Because many models pool audio tokens, we apply the same average-pooling operation to $a_{avg}$ to maintain alignment with the pooled audio indices, producing an importance map. 
Finally, we select the audio features with the highest attention scores ($\rho_a$\%) as the representative and information-dense tokens, while treating other tokens as non-significant.

\vspace{0.2em}
\noindent \textbf{Audio Anchor Consolidation.}
Considering the importance and pruning sensitivity of audio tokens, we merge a subset of non-salient tokens, thereby preserving semantic salience while maintaining context coverage. 
Specifically, for each time window, we uniformly sample anchors from the non-salient audio tokens. 
To maintain multimodal consistency, we evaluate candidates using cross-modal similarity between audio and video tokens:
\begin{equation}
\mathbf{S_{cross}} = \hat{H_{a}}\hat{H_{v}}^\top,
S_{ij} = \hat{h_{a}}_i^\top \hat{h_{v}}_j \in [-1,1],
\end{equation}
where $\hat{H_{a}}$, $\hat{H_{v}}$ denote the normalized audio token and video token sequences, respectively:
\begin{equation}
 \hat{H}
=\operatorname{Diag}\!\Big(\sqrt{\operatorname{diag}(HH^\top)}+\varepsilon\Big)^{-1}H, \varepsilon=10^{-6}.
\end{equation}
Then, we select the top-$\mathcal{G}$ audio tokens most related to the paired video segment and merge them into the anchor, where $\mathcal{G}$ is the number of merging tokens for each anchor.
Finally, the remaining non-salient tokens are discarded.

\input{table/tab_main}

\vspace{0.2em}
\noindent \textbf{Audio-Guided Video Token Compression.} 
In prior single-modal token-pruning work, it is hard to assess whether key information and events occur between frames \cite{tao2025dycoke,shen2025fastvid,shao2025holitom}. 
However, in OmniLLMs, introducing audio tokens is both challenging and beneficial. 
We set the total video token pruning ratio as $\rho_v$.
After filtering audio tokens, we map scores back to time windows and compute a per-window audio-retention score $S_a(i) \in [0,1]$, and $i$ is the index of the time group. 
Windows with high retention are deemed significant—providing information-dense, event-boundary cues. 
We dynamically prune video tokens: high-saliency windows are pruned conservatively, whereas low-saliency windows are pruned more aggressively.
Thus, we get the initial ratios $\rho'_{v}(i)$:
\begin{equation}
    \rho'_{v}(i) = \rho_{max} - (\rho_{max} - \rho_{min}) \cdot S_a(i),
\label{eq_dp}
\end{equation}
where $\rho_{max}$ and $\rho_{min}$ are the upper and lower limits of the pruning rate set to prevent excessive pruning.
These initial ratios $\rho'_{v}(i)$ are then algorithmically normalized to ensure the final rates $\rho_{v}$ strictly adhere to the global pruning budget. 
Overall, audio pruning remains more conservative, while video pruning is time-adaptive. 
This audio-guided strategy preserves key frames and temporal-alignment cues without additional training, while substantially reducing the total token count and inference overhead.

\subsection{ISTC Block}
\label{sec:2_istc}

In this section, we describe the interleaved spatio-temporal compression (ISTC) module used in OmniZip. 
Video token pruning is performed independently within each time window, and we set the minimum processing unit to four frames.
We interleave temporal-spatial redundancy evaluation for each frame and apply the corresponding strategies to compress tokens. 
As shown in \cref{fig-MED}, we first compute cosine similarity between same-position tokens in adjacent frames:
\begin{equation}
\mathbf{S_{vid}} =\cos(\theta)=\frac{h^{i}_v\cdot h^{j}_v}{\|h^{i}_v\|\|h^{j}_v\|},
\end{equation}
and use $\mathbf{S_{vid}}$ to estimate temporal redundancy and prune tokens in frames 2 and 4 with high similarity. For tokens in frames 1 and 3, we apply cluster-based pruning via density-peak clustering with k-nearest neighbors (DPC-KNN) \cite{du2016study}. For each video token $h^i_v$, we compute each token’s local density $\rho_i$ and its distance $\delta_i$ to the nearest higher-density token, yielding the final density score $\delta_i \times \rho_i$.
\begin{equation}
\rho_i
= \exp\!\left(
  -\frac{1}{k}\sum_{h^j_v \in \mathrm{kNN}(h^i_v)} d(h^i_v,h^j_v)^2
\right),
\end{equation}
\begin{equation}
\delta_i =
\begin{cases}
\displaystyle \max_{j\ne i} d(h^i_v,h^j_v), & \text{if } \rho_i = \max\limits_k \rho_k,\\[6pt]
\displaystyle \min_{j:\ \rho_j > \rho_i} d(h^i_v,h^j_v), & \text{otherwise.}
\end{cases}
,
\end{equation}
where $d(\cdot)$ is the euclidean distance. We prune tokens based on the density score, retaining salient video tokens and discarding spatially redundant ones.

\input{table/tab_worldsense}

\begin{figure*}[t]
    \centering
    \begin{subfigure}[t]{0.32\linewidth}
        \centering
        \captionsetup{font={footnotesize}, skip=4pt}
        \includegraphics[width=1\linewidth]{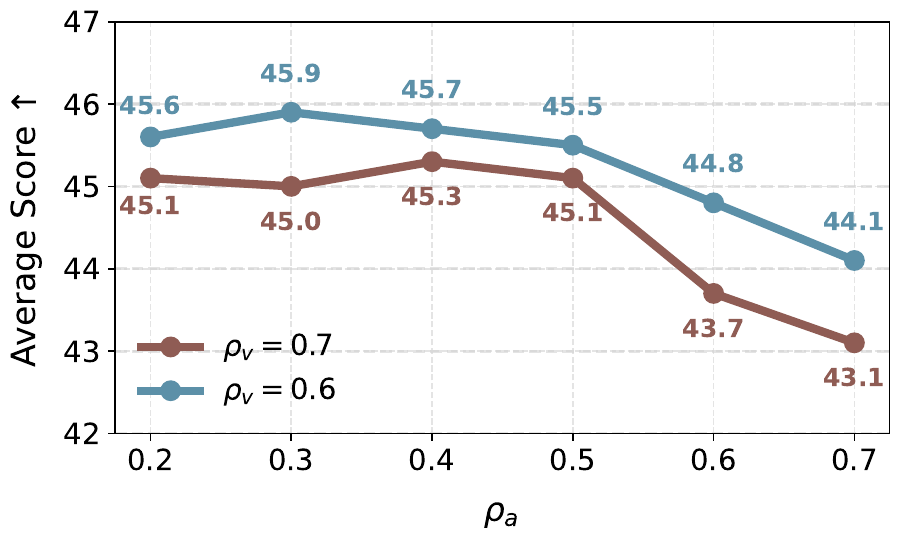}
    \end{subfigure}%
    \hspace{1mm}
    \begin{subfigure}[t]{0.32\linewidth}
        \centering
        \captionsetup{font={footnotesize}, skip=4pt}
        \includegraphics[width=1\linewidth]{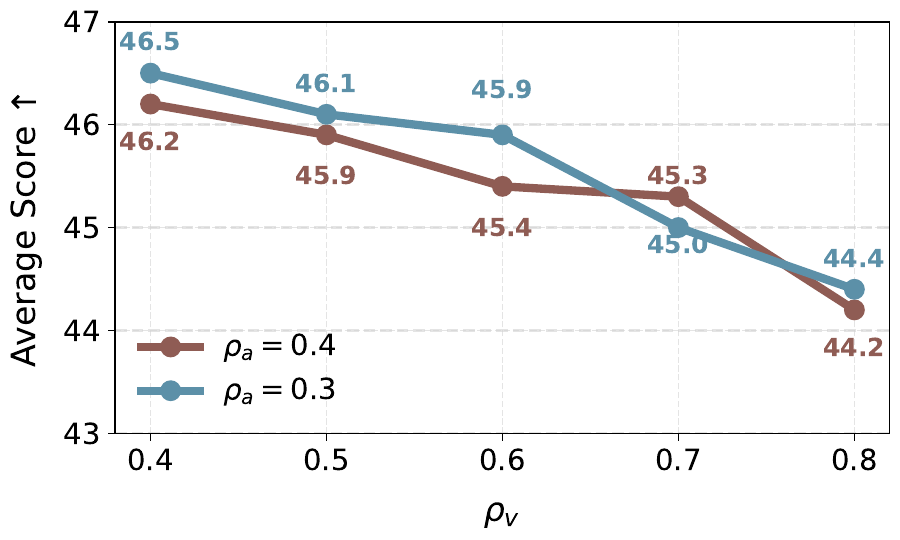}
    \end{subfigure}%
    \hspace{1mm}
    \begin{subfigure}[t]{0.32\linewidth}
        \centering
        \captionsetup{font={footnotesize}, skip=4pt}
        \includegraphics[width=1\linewidth]{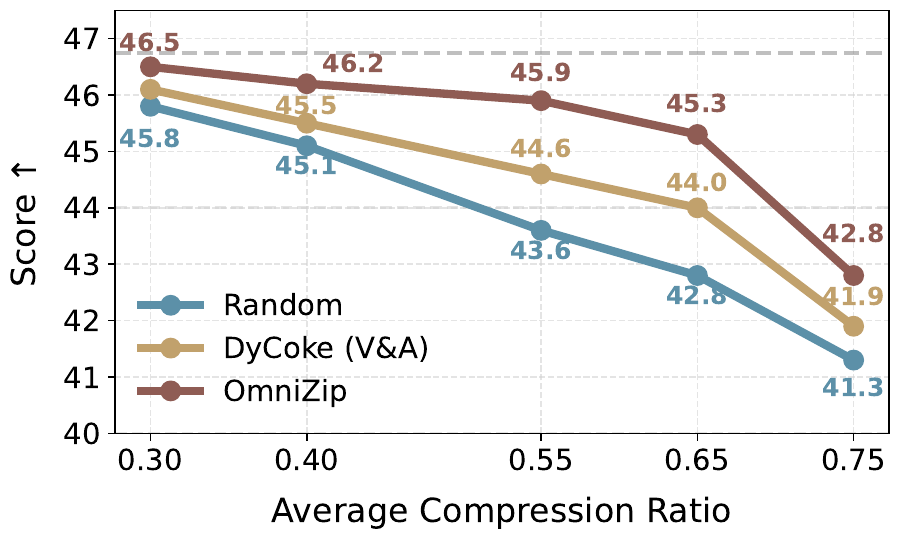}
    \end{subfigure}%
    \vspace{-4mm}
    \caption{\textbf{Ablation study on $\rho_a$ and $\rho_v$.} All experiments illustrated in the figure were carried out on the Qwen2.5-Omni-7B model and the WorldSense benchmark. \textbf{Left and Middle:} We separately analyze the influence of varying $\rho_a$ and $\rho_v$ on model performance. In general, excessive pruning of either modality negatively impacts model performance. However, an appropriate balance of audio and video token pruning achieves the best effect. \textbf{Right:} Performance of our method \emph{vs.} other methods in different compression ratios.}
    \label{fig:abs-rho}
\vspace{-2mm}
\end{figure*}

\subsection{Further Remarks on Our Method Design}
\label{sec:3_remark}
In this section, we analyzed the common limitations in prior work and further remark on our method design.
To our knowledge, OmniZip is the first token-compression framework for OmniLLMs in the audio–video understanding setting. 
In its design, we align with current developments in multimodal large language models and incorporate insights from prior work.
First, our method does not require accessing attention-score matrices inside the LLM, enabling compatibility with FlashAttention~\cite{dao2022flashattention,dao2023flashattention2} without incurring additional compute or memory overhead \cite{chen2024image,shao2025holitom,he2024zipvl}. 
It also preserves multi-round dialogue capability and remains compatible with other inference frameworks. 
Second, because most mainstream models now adopt ViT-based visual encoders, methods such as VisionZip can trigger GPU memory overflow when extracting attention-score matrices \cite{yang2025visionzip,shao2025holitom}; our approach avoids this issue. 
By contrast, the audio encoder is comparatively lightweight.
Finally, the additional runtime cost of token pruning is a common concern: OmniZip’s pruning step takes less than 40 ms, making it lightweight and not slowing inference.

%% file: table/tab_main.tex
\begin{table*}[t]
\centering
\captionsetup{skip=4pt}
\renewcommand{\arraystretch}{0.94} %
\setlength{\tabcolsep}{8pt} %
\resizebox{1\linewidth}{!}{
\begin{tabular}{ccccccccccccc}
\toprule
\multirow{2}{*}{Method} & \multicolumn{2}{c}{Settings} & \multicolumn{7}{c}{AVUTBench} & VideoMME & ShortVid-Bench & \multirow{2}{*}{Avg.} \\ \cmidrule(l){2-12} 
 & Retained Ratio & FLOPs Ratio & EL & OR & OM & IE & CC & CM & Avg. & wo & Avg. Score &  \\ \midrule
\multicolumn{13}{c}{\emph{Qwen2.5-Omni-7B}} \\ \midrule
\rowcolor{mycol2!40!white}
Full Tokens & 100\% & 100\% & 38.2 & 67.8 & 59.6 & 85.6 & 44.1 & 66.7 & 64.5 & 66.0 & 70.5 & 100\% \\ %
Random & 55\% & 48\% & 38.2 & 64.9 & 55.6 & 80.1 & 34.7 & \underline{65.0} & 61.0 & 65.4 & 68.3 & 96.9\% \\ %
FastV & 50\% & 54\% &  34.1 & 64.3 & \underline{57.1} & 77.6 & 36.4 & 56.4 & 58.4 & -  & 68.0  & 94.3\% \\ %
DyCoke (V\&A) & 50\% & 44\% & \textbf{38.8} & \textbf{67.2} & \textbf{58.2} & \underline{81.9} & \underline{39.0} & 62.4 & \underline{62.0} & \underline{65.5} & \underline{68.5} & 97.5\% \\ %
\rowcolor{mycol1}
OmniZip (Ours) & 45\% & 39\% & \underline{38.4} & \textbf{67.2} & 56.9 & \textbf{85.3} & \textbf{42.4} & \textbf{66.0} & \textbf{63.0} & \textbf{66.3} & \textbf{69.9} & \textbf{99.1\%} \\ %
\midrule
Random & 40\% & 34\% & 31.7 & 58.5 & 53.3 & 74.9 & 43.2 & \underline{59.0} & 56.9 & 65.0 & 67.7 & 94.3\% \\ %
FastV & 35\% & 42\% & 24.1 & 60.7 & 54.3 & \underline{81.6} & \underline{40.7} & 58.3 & \underline{57.8} & - & 67.9 & 93.8\%\\ %
DyCoke (V\&A) & 35\% & 29\% & \underline{32.9} & \underline{62.1} & \textbf{54.9} & 74.5 & 39.0 & 58.3 & 57.4 & \underline{65.2} & \underline{68.0} & 94.7\%  \\
\rowcolor{mycol1}
OmniZip (Ours) & 35\% & 29\% & \textbf{34.1} & \textbf{67.5} & \underline{54.6} & \textbf{83.7} & \textbf{42.4} & \textbf{61.2} & \textbf{61.0} &  \textbf{66.1} & \textbf{69.0} & \textbf{97.6\%} \\ %
\midrule
\multicolumn{13}{c}{\emph{Qwen2.5-Omni-3B}} \\ 
\midrule
\rowcolor{mycol2!40!white}
Full Tokens & 100\% & 100\% & 32.9 & 65.3 & 58.4 & 85.0 & 44.1 & 62.6 & 62.2 & 62.6 & 69.4 & 100\% \\
Random & 55\% & 45\% & 31.7 & 59.2 & 55.4 & 77.3 & \textbf{44.9} & \textbf{62.1} & 58.7 &  61.1 & 67.9 & 96.6\% \\
FastV & 50\% & 49\% & 27.1 & 57.0 & 56.3 & 80.5 & \underline{42.3} & 60.1 & 55.9 & - & \underline{68.0} & 95.7\% \\
DyCoke (V\&A) & 50\% & 40\% & \underline{31.9} & \underline{64.3} & \underline{57.3} & \underline{82.2} & 40.7 & 61.3 & \underline{60.7} & \underline{61.6} & 67.4 & 97.7\% \\
\rowcolor{mycol1}
OmniZip (Ours) & 45\% & 36\% & \textbf{32.4} & \textbf{65.0} & \textbf{57.7} & \textbf{84.9} & 41.5 & \underline{61.4} & \textbf{61.3} & \textbf{62.8} & \textbf{68.5} & \textbf{99.2\%} \\ 
\midrule
Random & 40\% & 31\% & \underline{28.2} & 60.8 & 54.9 & 73.1 & \underline{42.3} & \textbf{61.6} & 57.5 & 60.6 & 67.0 & 95.4\% \\
FastV & 35\% & 37\% & 24.2 & 60.8 & 54.3 & \underline{81.6} & 40.7 & 58.3 & \underline{57.7} & - & \underline{67.7} & 96.9\% \\
DyCoke (V\&A) & 35\% & 26\% & 32.9 & \underline{62.1} & 54.9 & 74.5 & 38.9 & 58.3 & 57.4 & \underline{61.0} & 67.5 & 95.7\% \\
\rowcolor{mycol1}
OmniZip (Ours) & 35\% & 26\% & \textbf{28.8} & \textbf{63.1} & \textbf{58.2} & \textbf{84.0} & \textbf{42.4} & \underline{60.4} & \textbf{60.1} & \textbf{62.7} & \textbf{68.0} & \textbf{98.3\%} \\
\bottomrule
\end{tabular}
}
\caption{\textbf{Comparison of different methods on omnimodal (audio \& video) QA benchmarks}. The \textbf{best} result among token pruning methods for each metric is in bold, and the \underline{second-best} is underlined. The `-' symbol indicates that FastV fails to execute due to an Out-of-Memory (OOM) error, and we also ignore its value when calculating the average score. The 'DyCoke (V\&A)' label denotes the application of its TTM module~\cite{tao2025dycoke} to both audio and video tokens.}
\label{tab:main-exp}
\vspace{-3mm}
\end{table*}

%% file: table/tab_worldsense.tex
\begin{table*}[t]
\centering
\captionsetup{skip=4pt}
\renewcommand{\arraystretch}{0.94} %
\setlength{\tabcolsep}{8pt} %
\resizebox{1\linewidth}{!}{
\begin{tabular}{cccccccccccc}
\toprule
Method & Retained Ratio & FLOPs (T) & \begin{tabular}[c]{@{}c@{}}Tech \&\\ Science\end{tabular} & \begin{tabular}[c]{@{}c@{}}Culture \&\\ Politics\end{tabular} & \begin{tabular}[c]{@{}c@{}}Daily\\ Life\end{tabular} & \begin{tabular}[c]{@{}c@{}}Film \&\\ TV\end{tabular} & Performance & Games & Sports & Music & Avg. \\ \midrule
\multicolumn{12}{c}{\emph{Qwen2.5-Omni-7B}} 
\\ 
\midrule
\rowcolor{mycol2!40!white}
Full Tokens & 100\% & 73.2 & 52.4 & 50.1 & 48.5 & 44.6 & 43.8 & 41.6 & 41.6 & 47.3 & 46.8 \\
Random & 55\% & 35.5 & 47.1 & 47.0 & 44.4 & 41.2 & 40.0 & 40.1 & 40.1 & 46.3 & 43.6 \\
FastV & 50\% & 39.3 & \underline{48.8} & 47.4 & 44.2 & 44.1 & \textbf{41.2} & 38.3 & 40.0 & \underline{46.6} & 44.3 \\
DyCoke (V\&A) & 50\% & 31.9 & 48.4 & \underline{49.9} & 46.7 & 41.4 & 39.9 & \textbf{40.8} & 40.2 & 46.5 & 44.6 \\
\rowcolor{mycol1}
OmniZip (Ours) & 45\% & 28.3 & \textbf{49.4} & \textbf{51.1} & \textbf{45.6} & \textbf{43.9} & \underline{40.1} & \textbf{40.8} & \underline{41.9} & \textbf{46.7} & \textbf{45.9} \\ 
\rowcolor{mycol1}
OmniZip (Ours) & 35\% & 21.4 & 48.3 & 49.5 & \textbf{47.6} & \underline{42.5} & \underline{40.1} & 40.2 & \textbf{42.3} & 46.3 & \underline{45.3} \\
\midrule
\multicolumn{12}{c}{\emph{Qwen2.5-Omni-3B}} 
\\ 
\midrule
\rowcolor{mycol2!40!white}
Full Tokens & 100\% & 37.4 & 51.5 & 50.8 & 45.0 & 45.4 & 43.8 & 42.5 & 44.2 & 46.1 & 46.4 \\
Random & 55\% & 17.0 & 48.2 & 46.3 & 40.7 & 41.4 & 38.6 & 40.0 & 41.8 & \textbf{43.4} & 42.8 \\
FastV & 50\% & 18.2 & \underline{50.0} & \textbf{50.5} & \textbf{44.1} & 43.0 & \textbf{40.5} & 41.6 & 41.8 & 42.1 & \underline{44.4} \\
DyCoke (V\&A) & 50\% & 15.1 & 48.1 & 48.5 & 42.3 & 43.3 & 39.7 & \textbf{43.4} & 42.1 & 43.0 & 44.0 \\
\rowcolor{mycol1}
OmniZip (Ours) & 45\% & 13.3 & \textbf{50.1} & \textbf{50.5} & \underline{43.9} & \underline{45.6} & \textbf{40.5} & 40.8 & \textbf{43.7} & \underline{43.1} & \textbf{45.2} \\
\rowcolor{mycol1}
OmniZip (Ours) & 35\% & 9.9 & 48.8 & 48.9 & 41.8 & \textbf{46.4} & 39.8 & \underline{42.5} & \underline{42.6} & \underline{43.1} & 44.3 \\
\bottomrule
\end{tabular}
}
\caption{\textbf{Comparison of different methods on the WorldSense benchmark}. The \textbf{best} result among token pruning methods for each metric is in bold, and the \underline{second-best} is underlined. The FLOPs calculation considers only the multimodal tokens originating from audio and video inputs. FastV failed to run on the 7B model due to an OOM error on an A6000 GPU, so we evaluated its performance on a single H100 (80G) GPU.
}
\label{tab:WorldSense}
\vspace{-3mm}
\end{table*}

%% file: sec/4_exp.tex
\section{Experimental Results}
\label{sec:exp}

\subsection{Evaluation Setups and Implementation Details}
\vspace{0.2em}
\textbf{Benchmarks.} We evaluate the performance of OmniLLMs using established audio-video understanding benchmarks: AVUT~\cite{yang2025audio}, VideoMME~\cite{fu2025video}, ShortVid-Bench~\cite{ge2025arc}, and WorldSense~\cite{hong2025worldsense}. 
Among these benchmarks, VideoMME is widely used for pure video-understanding evaluations, and including audio can improve accuracy. 
AVUT is an \emph{audio-centric} video understanding benchmark focusing on six tasks: event localization (EL), object matching (OM), OCR matching (OR), information extraction (IE), content counting (CC), and character matching (CM). 
WorldSense assesses models’ ability to understand over audio and video across eight domains jointly. 
ShortVid-Bench evaluates the ability of models to understand real-world short videos.

\vspace{0.2em}
\noindent \textbf{Comparison Methods.} 
Given the absence of token pruning methods specifically designed for the omnimodal setting, we select representative prior methods from single-modal domains for adaptation and comparative analyses. 
FastV~\cite{chen2024image}, during its prefill stage, utilizes the attention score matrix of the $L$-th layer to evaluate token relevance, subsequently pruning tokens. 
DyCoke~\cite{tao2025dycoke} represents the first dynamic token compression strategy proposed for VideoLLMs. 
We employ its first-stage TTM module to process video and audio tokens. 
Furthermore, we implement a random pruning as a control group to provide a rigorous comparative analysis.

\vspace{0.2em}
\noindent \textbf{Implementation Details.}
We implement the proposed OmniZip on the Qwen2.5-Omni (7B and 3B) models using NVIDIA A6000 (48GB) GPUs~\cite{xu2025qwen2}.
To set pruning ratios across methods, we use the overall FLOPs ratio as the metric to ensure a fair comparison. 
For FastV, we set the attention-computation layer to layer 5.
For video input, to better match the time-window granularity—and given that VideoMME videos are relatively long—we cap the maximum number of frames at 768. 
For other datasets, we cap inputs at 128 frames. 
For each time window, it has 50 audio tokens and 288 video tokens.
For hyperparameter settings, we set $\rho_{max} = 0.75$, $\rho_{min} = 0.35$, $k=5$ and $\mathcal{G}=15$ for AVUT and $\mathcal{G}=3$ for others.
For 45\% and 35\% retained ratio, we set $\rho_a=0.3, \rho_v=0.6$ and $\rho_a=0.4, \rho_v=0.7$ respectively, except for ShortVid-Bench. 
For all experiments, we leverage FlashAttention to reduce memory usage.

\subsection{Main Results}
We evaluate our approach on recent mainstream models Qwen2.5-Omni at two parameter scales (7B and 3B). 
For the VideoMME, we use the LMMs-Eval \cite{zhang2024lmmsevalrealitycheckevaluation, lmms_eval2024} for evaluation, and for other benchmarks, we follow the unified testing code for all experimental settings. 
We evaluated performance and inference cost at two distinct token retention rates.
To facilitate a comprehensive evaluation, the results in \cref{tab:main-exp} are normalized and presented as percentages, where the baseline model's accuracy is set to 100\%. 
Notably, unlike conventional purely video understanding tasks, audio-video understanding tasks present greater \emph{challenges} and exhibit \emph{increased sensitivity} to token pruning.

\begin{figure}[t]
\centering
\captionsetup{font={small}, skip=1pt}
\includegraphics[width=1\linewidth]{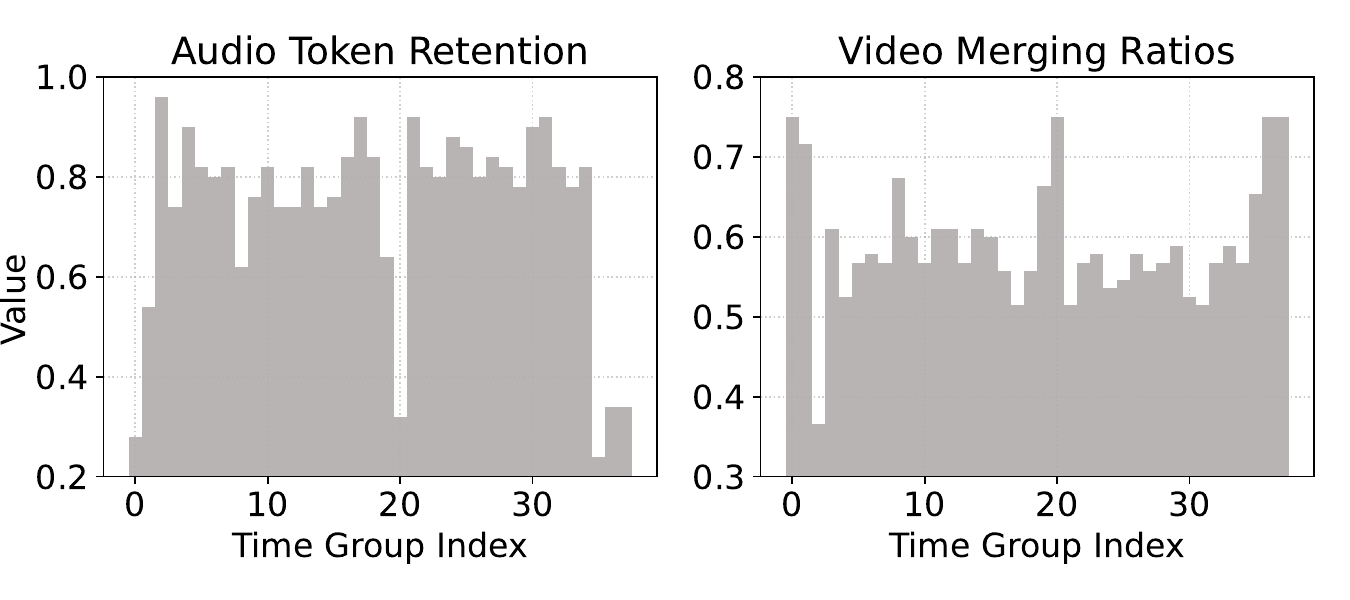}
\vspace{-5mm}
\caption{\textbf{Visualization of dynamic pruning ratios.} The figure illustrates how audio token retention guides the allocation of video token pruning. Specifically, for time windows with low audio retention, we allocate a higher video pruning ratio, while maintaining a constant total pruning rate.}
\label{fig:ratio_V}
\vspace{-4mm}
\end{figure}
\vspace{0.2em}
\noindent \textbf{Comparison with State-of-the-Art Methods.} As shown in \cref{tab:main-exp}, the results indicate that OmniZip maintains optimal performance with the fewest tokens across diverse test benchmarks. Even with a 60\% reduction in computational FLOPs, the model retains an average accuracy of 99.1\%. 
In contrast, the random pruning leads to significant performance degradation. 
FastV similarly fails to achieve effective results, a limitation attributable to the uneven attention distribution between video and audio tokens and the consequent disruption of temporal windows.
DyCoke is designed to reduce redundancy in the temporal dimension while preserving the time window structure. 
However, as it is designed for single-modal video and neglects spatial redundancy, its omnimodal performance is suboptimal. 
At lower retention rates, OmniZip maintains its leading performance.
Besides, as shown in \cref{tab:WorldSense} for the WorldSense Benchmark, OmniZip at a 35\% token retention rate outperforms other methods operating at a 50\% retention rate.

Furthermore, our experiments across different model scales reveal that models with fewer parameters are more amenable to compression, corroborating prior studies \cite{tao2025dycoke,shao2025holitom}. 
We also note that the missing FastV results for the 7B model are attributable to its incompatibility with Flash Attention, which requires the explicit calculation of the attention matrix and subsequently causes an out-of-memory (OOM) error. 
We circumvent this problem in our method design.

\input{table/time_eval}

\vspace{0.2em}
\noindent \textbf{Sensitivity Analyses on $\rho_a$ and $\rho_v$.} 
As illustrated in the left and middle plots of \cref{fig:abs-rho}, excessive pruning of either audio or video significantly degrades model performance. 
This suggests that due to the varying redundancy and attention given to audio and video tokens, identifying an optimal pruning ratio is crucial for maximizing compression effectiveness, a conclusion supported by the data. 
Thus, we suggest that the pruning rate can be dynamically adapted based on the specificity of the task, such as its relative dependence on video or audio information.
Moreover, our results suggest that the audio token pruning rate should be lower than the video token pruning rate. 
Finally, the right plot indicates that OmniZip outperforms other methods across all pruning ratios. As the pruning rate increases, our accuracy declines more gradually, highlighting the robustness of our method.

\begin{figure}[t]
\centering
\captionsetup{font={small}, skip=2pt}
\includegraphics[width=1\linewidth]{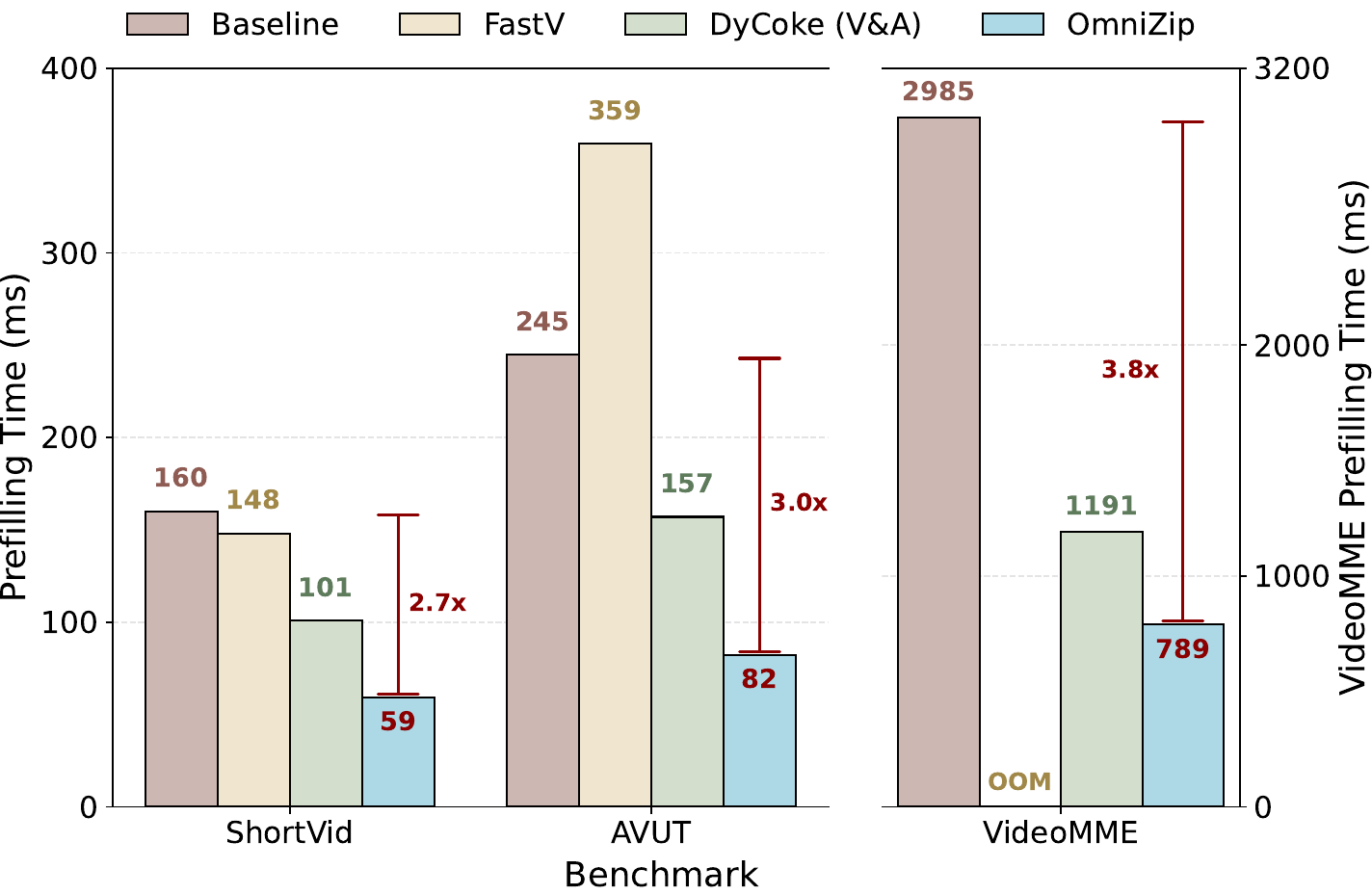}
\caption{\textbf{Achieving superior inference speedup.} We visualize the inference speedup achieved by OmniZip during the prefilling stage on the 7B model. As video sequence length increases, the speedup effect becomes more pronounced. OmniZip achieves a 2.7–3.8$\times$ inference speedup while robustly maintaining model accuracy. }
\label{fig:speed_analy}
\vspace{-5mm}
\end{figure}

\input{table/abs_select_method}

\vspace{0.2em}
\noindent \textbf{Visualization of Dynamic Pruning.} 
\cref{fig:ratio_V} visualizes the dynamic allocation of pruning rates in OmniZip, illustrating that across different time windows, the pruning rate of video tokens changes dynamically in conjunction with that of audio tokens. 
Our method employs a dynamic pruning rate while simultaneously maintaining a constant overall pruning rate, which facilitates a fair comparison against other methods.
Collectively, this finding demonstrates the efficacy of our proposed approach; it also underscores the necessity of developing specialized research for the OmniLLMs.

\subsection{Efficiency Analyses}
We evaluated the inference speed and memory consumption across four benchmarks. 
As shown in \cref{tab:exp-mem-speed}, we conducted more detailed analyses on the WorldSense benchmark.
The results indicated that our method significantly accelerated inference speed compared to the full token model. 
On the 3B model, our method achieves 3.27$\times$ speedup in the prefilling stage. This advantage became more pronounced for larger models (7B), yielding a 1.42$\times$ speedup in overall inference and a 3.42$\times$ speedup in prefilling.
Moreover, our method significantly reduces the memory cost during inference. While maintaining an accuracy of approximately 97\%, the method reduces memory consumption by \emph{10G}, which is crucial for the practical deployment of OmniLLMs.

Furthermore, \cref{fig:speed_analy} summarizes the inference speedup on other benchmarks. 
Our method significantly reduces the prefilling stage time. 
In contrast, FastV is incompatible with Flash Attention due to its requirement for explicit attention matrix computation, which incurs extra overhead and consequently slows inference. 
Furthermore, due to inherent dataset characteristics (i.e., ShortVid comprises shorter videos while VideoMME features longer ones), the speedup on VideoMME is correspondingly more pronounced. 
Compared to the baseline, OmniZip achieves a 2.7–3.8$\times$ inference speedup, the highest among all methods.

\subsection{Ablation Study}

\vspace{0.2em}
\noindent \textbf{Ablation Study of DP and AC Technology.}
\cref{tab:abd_omnizip} presents an ablation study on the two core components of the OmniZip framework: dynamic video pruning (DP) and audio anchor consolidation (AC). As shown in the table, removing the dynamic pruning allocation for video tokens significantly decreases model accuracy. Further eliminating the audio anchor consolidation strategy leads to an additional performance degradation. This result validates the efficacy and design rationale of OmniZip.

\vspace{0.2em}
\noindent \textbf{Ablation Study about Token Selection Method.}
\cref{tab:abs_selection_method} presents a comparative analysis of different token selection strategies. 
First, \cref{tab:abs_selection_method} (ID:2) demonstrates the superior performance of ISTC over DyCoke for video tokens. 
Furthermore, VisionZip~\cite{yang2025visionzip}, a global token selection (GS) strategy, is included in the comparison. The results indicate that the GS strategy is suboptimal for the omnimodal setting. 
The GS strategy extracts focused video and audio tokens independently, ignoring semantic alignment and disrupting the temporal structure, making it difficult to maintain model accuracy.
Notably, the additional computation required by VisionZip to compute the visual attention matrix frequently causes OOM, a limitation that OmniZip avoids.
Besides, a comparison against random selection underscores the effectiveness of our method.
Therefore, our method represents a specialized design that accounts for the characteristics of multimodal information, offering clear advantages over prior single-modal token compression methods.

\section{Conclusion}
This paper presents \textit{OmniZip}, a novel \textit{training-free} method to dynamically reduce the audio-video tokens based on audio-guidance for faster omnimodal large language models (OmniLLMs). 
Specifically, the framework first identifies salient audio tokens and calculates an audio retention rate for each time window, which is then used to dynamically guide the pruning of video tokens in conjunction with a corresponding spatio-temporal compression module. To the best of our knowledge, this is the first token pruning method tailored to OmniLLMs that jointly optimizes the compression of multimodal audio-video tokens.
Extensive benchmark and analysis results on a wide range of audio-video understanding tasks with two OmniLLMs (3B, 7B parameters) demonstrate that our method consistently surpasses prior single-modal methods.
Our method achieves up to a 10G memory reduction and a 2.7–3.8$\times$ prefill speedup, while maintaining nearly identical performance.

%% file: table/time_eval.tex
\begin{table}[t]
\captionsetup{font={small}, skip=3pt}
\centering
\setlength{\tabcolsep}{4pt}
\resizebox{1.0\linewidth}{!}{
\begin{tabular}{lcccc}
\toprule
\multicolumn{1}{l}{Method} & GPU Mem. $\downarrow$ & Prefilling Time $\downarrow$& Acc. $\uparrow$& Latency per Example $\downarrow$\\ \midrule
\multicolumn{5}{c}{Qwen2.5-Omni-7B} \\ \midrule
\rowcolor{mycol2!40!white}
Full Tokens & 35G & 291ms (1.00$\times$)& 46.8 & 4.52s (1.00$\times$) \\
\rowcolor{white}
FastV & \multicolumn{4}{c}{OOM} \\
DyCoke (V\&A) & 31G & 184ms (1.58$\times$)& 44.6 & 3.64s (1.24$\times$) \\
\rowcolor{mycol1}
Ours (45\%) & 28G & 116ms (2.51$\times$)& 45.9 & 3.40s (1.33$\times$) \\ 
\rowcolor{mycol1}
Ours (35\%) & \textbf{25G} & \textbf{85ms (3.42$\times$)} & 45.3 & \textbf{3.18s (1.42$\times$)} \\ 
\midrule
\multicolumn{5}{c}{Qwen2.5-Omni-3B} \\ 
\midrule
\rowcolor{mycol2!40!white}
Full Tokens & 25G & 258ms (1.00$\times$)& 46.4 & 3.61s (1.00$\times$)\\
\rowcolor{white}
FastV & 45G & 222ms (1.16$\times$)& 44.4 & 3.45s (1.05$\times$)\\
DyCoke (V\&A) & 20G & 171ms (1.51$\times$)& 44.0 & 3.12s (1.16$\times$)\\
\rowcolor{mycol1}
Ours (45\%) & 17G & 104ms (2.48$\times$)& 45.2 & 2.86s (1.26$\times$)\\ 
\rowcolor{mycol1}
Ours (35\%) & \textbf{16G} & \textbf{79ms (3.27$\times$)} & 44.3 & \textbf{2.75s (1.31$\times$)}\\ 
\bottomrule
\\
\end{tabular}
}
\vspace{-2.mm}
\caption{\textbf{Actual inference efficiency comparison on WorldSense}. Experiments with the 7B and 3B models are conducted on a single A6000 GPU. FastV computes the full attention matrix in memory, a process that results in Out-of-Memory (OOM) errors attributable to the large number of tokens. Our method can achieve the best model performance, the lowest memory consumption, and the greatest inference acceleration.}
\label{tab:exp-mem-speed}
\vspace{-3mm}
\end{table}

%% file: table/abs_select_method.tex
\begin{table*}[t]

\begin{minipage}{0.54\textwidth}
\centering
\scriptsize
\resizebox{1\linewidth}{!}{
\begin{tabular}{ccccccccccccc}
\toprule
\multirow{2}{*}{ID} & \multicolumn{3}{c}{Select Method} & \multicolumn{1}{c}{AVUT} & WorldSense & ShortVid-Bench \\ \cmidrule(l){2-13} 
 & Video & Audio & GS & Avg. & wo & Avg. Score \\ 
\midrule
\rowcolor{mycol2!40!white}
Baseline & - & - & - & 64.5 & 46.8 & 70.5  \\
\textbf{\emph{1}} & ISTC & Random & \ding{55} & 60.0 & 45.1 & 69.0 \\
\textbf{\emph{2}} & DyCoke & Ours & \ding{55} & 62.1 & 45.0 & 69.2 \\
\textbf{\emph{3}} & VisionZip & Ours & \ding{51} & 61.4 & 44.2 & 68.0 \\
\textbf{\emph{4}} & Random & Random & \ding{51} & 60.4 & 43.3 & 68.1 \\
\rowcolor{mycol1}
OmniZip & ISTC & Ours & \ding{55} & \textbf{63.0} & \textbf{45.9} & \textbf{69.9}  \\
\bottomrule
\end{tabular}
}
\vspace{-2.4mm}
\captionsetup{font={small}}
\caption{\textbf{Ablation study of the token selection method.} We compare our token selection method against baseline strategies on 7B model. Furthermore, to substantiate the design rationale of OmniZip, we compare it against VisionZip~\cite{yang2025visionzip}, a method that performs global video token selection (GS).}
\label{tab:abs_selection_method}
\end{minipage}%
\hspace{00.5mm} 
\begin{minipage}{0.44\textwidth}
\vspace{-0.23mm}
\resizebox{1\linewidth}{!}{
\begin{tabular}{ccccccc}
\toprule
\multicolumn{3}{c}{Settings} & AVUT & WorldSense & ShortVid-Bench \\ \midrule
Re. Ratio & DP & AC & Avg. & Avg. & Avg. \\ \midrule
\rowcolor{mycol2!40!white}
100\% & - & - & 64.5 & 46.8 & 70.5 \\
45\% & \ding{51} & \ding{51} & 63.0 & 45.9 & 69.9 \\
45\% & \ding{55} & \ding{51} & 62.0 \red{(-1.0)} & 45.0 \red{(-0.9)} & 69.3 \red{(-0.6)} \\
45\% & \ding{55} & \ding{55} & 61.7 \red{(-1.3)} &  44.8 \red{(-1.1)} &  69.0 \red{(-0.9)} \\ 
\bottomrule
\end{tabular}
}
\vspace{-1.5mm}
\captionsetup{font={small}}
\caption{\textbf{Ablation study of DP \& AC Technology.} To validate the efficacy of our method, we conduct an ablation study evaluating the impact of our two key components on final model accuracy: audio-guided dynamic video pruning (DP) and audio anchor consolidation (AC) on Qwen2.5-Omni-7B.
}
\label{tab:abd_omnizip}
\end{minipage}
\vspace{-3mm}
\end{table*}

%% file: sec/X_suppl.tex
\appendix
\clearpage
\setcounter{page}{1}
\maketitlesupplementary

\section{Dynamic Pruning Rate Allocation Algorithm}
\label{sec:dp_alg}

This section expands upon the audio-guided video token compression algorithm described in \cref{sec:2_omnizip}. \cref{alg:adaptive_pruning} defines the calculation for the dynamic pruning rate and illustrates that while this rate is adaptive, the overall pruning rate remains constant.

\begin{algorithm}[H]
\caption{Audio-guided Video Token Pruning}
\label{alg:adaptive_pruning}
\begin{algorithmic}[1] %
    \STATE \algorithmicparameter $\rho_{\min}$, $\rho_{\max}$, $\rho_v$
    \STATE \algorithmicrequire Audio-retention ratio $S_a = [S_a(1), ..., S_a(N)]$
    \STATE \algorithmicensure DP rates $\rho_v' = [\rho_v'(1), ..., \rho_v'(N)]$
    
    \STATE $N \gets \text{length}(S_a)$
    \STATE $\rho'_{v\_initial} \gets []$
    
    \STATE \COMMENT{Step 1: Compute initial pruning ratios (\cref{eq_dp})}
    \FOR{$i \gets 1$ \textbf{to} $N$}
        \STATE $\rho'_v(i) \gets \rho_{\max} - (\rho_{\max} - \rho_{\min}) \cdot S_a(i)$
        \STATE $\rho'_{v\_initial}.\text{append}(\rho'_v(i))$
    \ENDFOR
    
    \STATE \COMMENT{Step 2: Normalize to meet the global budget}
    \STATE $T_{budget} \gets \rho_v \times N$
    \STATE $T_{initial} \gets \sum(\rho'_{v\_initial})$
    \STATE $\rho_v' \gets \text{NormalizeRatios}(\rho'_{v\_initial}, T_{initial}, T_{budget})$
    
    \STATE \textbf{return} $\rho_v'$
    \STATE \textbf{end function}
\end{algorithmic}
\end{algorithm}

\section{Discussion}

\subsection{Adaptivity of OmniZip}
The design of OmniZip is motivated by an analysis of audio-visual tokens and the dominant paradigm of their time-window-based arrangement in OmniLLMs.
Notably, current mainstream models are generally based on this time-window paradigm~\cite{xu2025qwen2,xu2025qwen3,yang2025humanomniv2,ye2025omnivinci,tang2025video,fu2025vita}. This approach divides the continuous audio-visual stream into discrete time segments, fuses or concatenates the tokens from each modality within their respective segments, and finally inputs the combined sequence into a large language model.
This architectural commonality facilitates the adaptation of OmniZip to other existing models.

We also acknowledge that the field of OmniLLMs is still nascent, which raises the reasonable question of whether OmniZip would lose efficacy if some models no longer rely on explicit time-window concatenation. 
We argue that the core principle of OmniZip exploits the inherent temporal locality of audio-visual data streams. Within any short time segment, there is a high degree of correlation and synchronization between audio and video, accompanied by significant redundancy. 
Therefore, OmniZip remains a viable strategy, as its core mechanism—guiding token pruning by analyzing multi-modal tokens within a local temporal window—is fundamentally feasible and effective.

\subsection{Hardness of Omnimodal Token Compression}
While prior work in visual token compression has achieved high reduction rates (e.g., 70-85\%), this is because a single modality is inherently simpler to compress. 
However, for OmniLLMs, the variable contribution of audio and video across different tasks, and the fact that audio information, as a high-dimensional feature, is less intuitively compressible than visual data, complicates this process. 
Additionally, recent models increasingly incorporate token efficiency as a core design principle, making further gains from simple pruning more difficult to achieve. 
Therefore, token pruning audio-video tokens is significantly more challenging. 
Nevertheless, achieving comprehensive video understanding necessitates the joint processing of both audio and visual information, making an effective token compression strategy all the more critical. 
In summary, as the first audio-visual token compression method, OmniZip sets a new benchmark for future technological advancements.

\section{Computing Cost Evaluation}
We examine the total FLOPs introduced by \emph{audio tokens} and \emph{video tokens} of the prefilling stage and the decoding stage. 
In OmniLLMs, a transformer layer comprising a multi-head attention (MHA) module and a feed-forward network (FFN) module is considered. Here, $n$ denotes the token count, $d$ the hidden state dimension, and $m$ the FFN intermediate dimension.
In the prefilling phase, the total FLOPs can be approximated as  $4nd^2 + 2n^2d + 2ndm$. 
In the decoding phase, taking into account the significant contribution introduced by the KV cache the computational consumption for $\mathcal{R}$ total iterations (\textit{i.e.}, predicting $\mathcal{R}$ tokens) is $\mathcal{R}\left(4d^2+2dm\right)+2\sum_{i=1}^{\mathcal{R}}d\times(n+i)$. 
We unify $\mathcal{R}=100$ for calculation in the experiments. 
Thus, for an LLM with $T$ total transformer layers, the total FLOPs can be expressed as follows,
\begin{equation}
\begin{aligned}
\mathrm{FLOPs}&=T(4nd^2+2n^2d+2ndm)\\
&+T\mathcal{R}\left((4d^2+2dm)+2\left(dn+\frac{d(\mathcal{R}+1)}{2}\right)\right).
\end{aligned}
\end{equation}

\begin{table*}[t]
\centering
\captionsetup{skip=4pt}
\renewcommand{\arraystretch}{0.94} %
\setlength{\tabcolsep}{8pt} %
\resizebox{1\linewidth}{!}{
\begin{tabular}{ccccccccccccc}
\toprule
\multirow{2}{*}{Method} & \multicolumn{3}{c}{Settings} & \multirow{2}{*}{\begin{tabular}[c]{@{}c@{}}Tech \&\\ Science\end{tabular}} & \multirow{2}{*}{\begin{tabular}[c]{@{}c@{}}Culture \&\\ Politics\end{tabular}} & \multirow{2}{*}{\begin{tabular}[c]{@{}c@{}}Daily\\ Life\end{tabular}} & \multirow{2}{*}{\begin{tabular}[c]{@{}c@{}}Film \&\\ TV\end{tabular}} & \multirow{2}{*}{Performance} & \multirow{2}{*}{Games} & \multirow{2}{*}{Sports} & \multirow{2}{*}{Music} & \multirow{2}{*}{Avg.} \\ 
\cmidrule(l){2-4}  & Retained Ratio & $\rho_a$ & $\rho_v$ & & & & & \\
\midrule
\multicolumn{12}{c}{\emph{Qwen2.5-Omni-7B}} 
\\ 
\midrule
\rowcolor{mycol2!40!white}
Full Tokens & 100\% & - & - & 52.4 & 50.1 & 48.5 & 44.6 & 43.8 & 41.6 & 41.6 & 47.3 & 46.8 \\
Random & 55\% & 0.45 & 0.45 & 47.1 & 47.0 & 44.4 & 41.2 & 40.0 & 40.1 & 40.1 & 46.3 & 43.6 \\
FastV & 50\% & 0.5 & 0.5 & 48.8 & 47.4 & 44.2 & 44.1 & 41.2 & 38.3 & 40.0 & 46.6 & 44.3 \\
DyCoke (V\&A) & 50\% & 0.5 & 0.5 & 48.4 & 49.9 & 46.7 & 41.4 & 39.9 & 40.8 & 40.2 & 46.5 & 44.6 \\
\rowcolor{mycol1}
OmniZip (Ours) & 50\% & 0.5 & 0.5 & 50.4 & 49.5 & 47.7 & 42.5 & 41.6 & 41.2 & 42.8 & 47.8 & 46.1 \\
\midrule
DyCoke (V\&A) & 45\% & 0.55 & 0.55 & 47.1 & 49.5 & 44.5 & 41.2 & 40.8 & 40.7 & 40.5 & 46.6 & 44.1 \\
\rowcolor{mycol1}
OmniZip (Ours) & 45\% & 0.55 & 0.55 & 50.0 & 49.8 & 47.6 & 42.7 & 40.1 & 40.7 & 41.2 & 47.8 & 45.5 \\
\rowcolor{mycol1}
OmniZip (Ours) & 45\% & 0.3 & 0.6 & 50.1 & 51.1 & 47.6 & 43.9 & 40.1 & 40.8 & 41.9 & 46.7 & 45.9 \\ 
\bottomrule

\end{tabular}
}
\caption{\textbf{Comparison of different methods on the WorldSense benchmark}. FastV failed to run on the 7B model due to an OOM error on an A6000 GPU, so we evaluated its performance on a single H100 (80G) GPU. $\rho_a$ and $\rho_v$ are the pruning ratios of audio tokens and video tokens, respectively. 
}
\label{tab:WorldSense-appendix}
\vspace{-3mm}
\end{table*}

\section{Related Work}

\subsection{Video Large Language Models}
Video large language models (VideoLLMs) extend traditional LLMs and visual-language models~\cite{liu2023visual,feng2025can,feng2025rewardmap}, integrating video and language understanding into a unified framework~\cite{li2023videochat,lin2023video,liu2023visual,zhang2023video,bai2025qwen2,zhang2025videollama,wang2024qwen2,chen2024internvl,team2025gemma,llava-ov,an2025llava}. By jointly processing text and video inputs, VideoLLMs can perform complex cross-modal reasoning tasks, such as visual question answering and video captioning. 
They typically utilize pre-trained visual encoders and leverage powerful language backbones to align heterogeneous representations in a shared semantic space. 
Recent advancements, such as Qwen3-VL~\cite{qwen3technicalreport}, InternVL3.5~\cite{wang2025internvl3}, and Kimi-VL~\cite{team2025kimi}, have significantly advanced video-text understanding capabilities. 
However, as video inherently contains both visual and audio information, audio-video understanding is a key future research direction.

\subsection{Omnimodal Large Language Models}
To achieve a more human-like multimodal interaction experience, OmniLLMs have emerged. By leveraging multimodal data, they learn richer contextual information and achieve a deeper understanding of inter-modal relationships~\cite{xu2025qwen2,tong2025interactiveomni,xu2025qwen3,xie2024mini,tang2025video,zhang2024videoin,yang2025humanomniv2,ge2025arc,fu2024vita,shu2025audio,sun2024video,li2024baichuanomni,tao2026lvomnibench}. 
In video understanding tasks, compared to VideoLLMs, OmniLLMs can additionally consider audio information alongside visual data, enabling more realistic answers and a more comprehensive understanding. 
Recent work, such as Qwen2.5-Omni~\cite{xu2025qwen2}, introduced an end-to-end model capable of perceiving all modalities. 
While InteractiveOmni~\cite{tong2025interactiveomni} has enabled multi-round audio-video conversations, significant recent work~\cite{ai2025ming,xu2025qwen3,ye2025omnivinci,yang2025humanomniv2} has further advanced state-of-the-art omnimodal understanding capabilities.
However, the large number of multimodal tokens introduced by video and audio inputs significantly impedes the practical deployment and application of OmniLLMs. 
Balancing model performance and computational efficiency remains a significant challenge. 
Thus, developing efficient methods to simplify the token input derived from audio-video tokens is essential.

\subsection{Token Compression}
Recent research has focused on token compression to enhance the inference efficiency of multimodal large language models. 
This approach is highly effective as multimodal inputs often contain significant redundancies, such as image~\cite{bolya2022token,shang2024llava,yang2025visionzip,chen2024image,xing2024pyramiddrop,ye2025fit,yang2025topv,tan2025tokencarve,feng2025efficient,jin2025mergemix}, video~\cite{chen2025streamingtom,tao2025dycoke,shao2025holitom,huang2024prunevid,shen2025fastvid,shen2024longvu,ye2025fit,liu2025video}, and audio~\cite{li2023accelerating,lin2024speechprune,lee2025token,sun2024video}. 
A key advantage is that these methods can be applied as a tuning-free, post-processing technique. 
These methods operate by first establishing a metric to evaluate token importance, followed by corresponding compression operations~\cite{shao2025tokens}. 
While token compression methods for single modalities have been widely studied, their application to the omnimodal setting has not yet been explored. 
Furthermore, current mainstream methods typically depend on accessing the attention matrices from either the video encoder or the LLM~\cite{tao2025dycoke,shao2025holitom,yang2025visionzip,he2024zipvl,xing2024pyramiddrop}. 
This dependency is often incompatible with modern optimizations such as FlashAttention~\cite{dao2022flashattention,dao2023flashattention2}, necessitating the materialization of the full attention matrix. 
In conjunction with ultra-long visual token sequences, this readily leads to Out-of-Memory (OOM) errors. Therefore, such methods exhibit poor scalability to larger, more advanced models.    
Considering the inherent coupling of video and audio, we conduct the first exploration of token compression for the combined audio-video understanding task, aiming to facilitate the practical deployment of OmniLLMs.

\begin{figure*}[t]

\captionsetup{font={small}, skip=2pt}
\scriptsize
\centering
\resizebox{1\linewidth}{!}{
\begin{tabular}{ccc}
\hspace{-0.5cm}
\begin{adjustbox}{valign=t}
\begin{tabular}{c}
\end{tabular}
\end{adjustbox}
\begin{adjustbox}{valign=t}
\begin{tabular}{cccccc}
\includegraphics[width=0.49\linewidth]{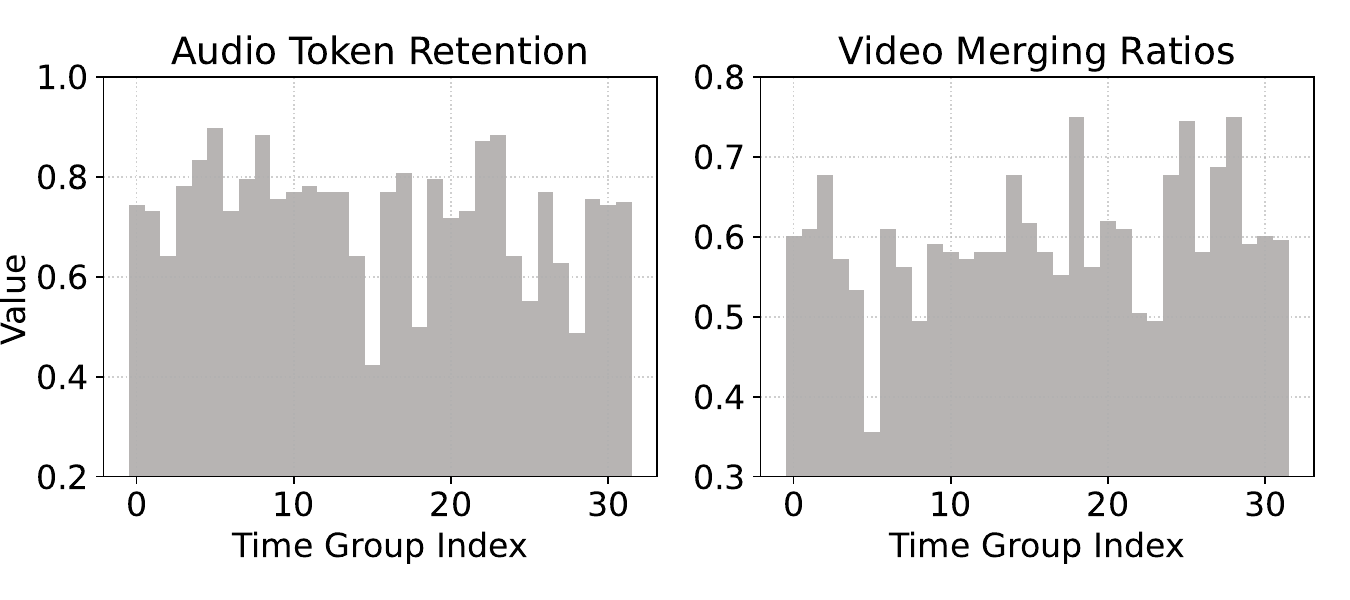} \hspace{-4mm} &
\includegraphics[width=0.49\linewidth]{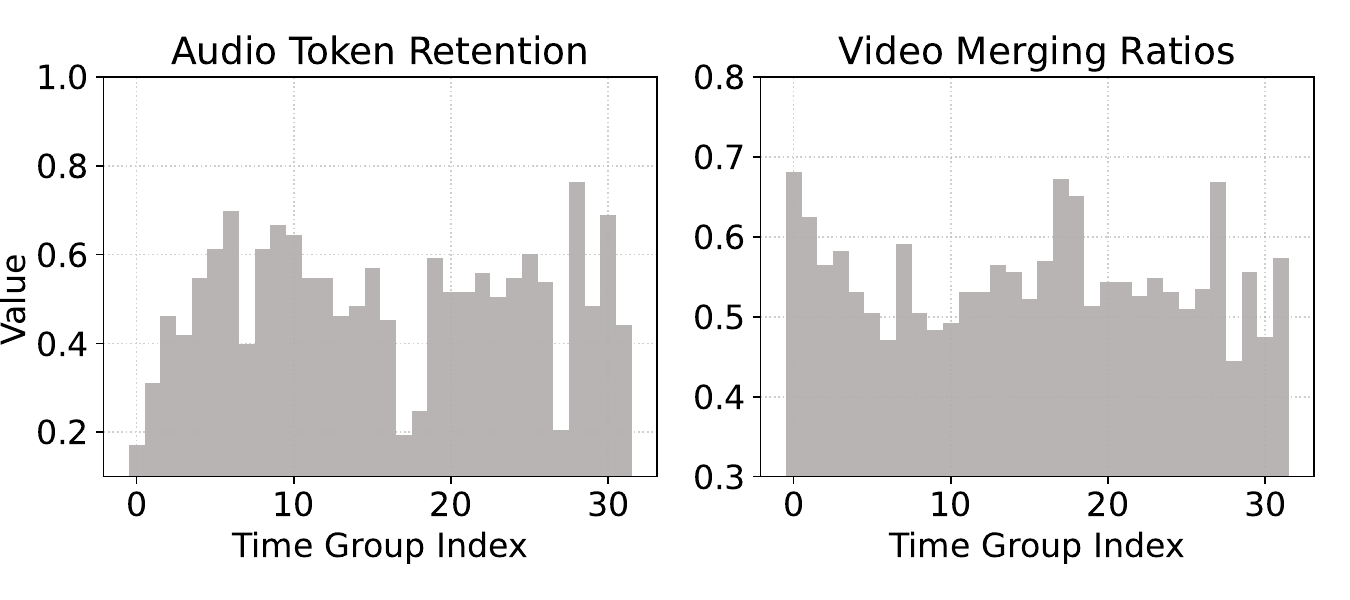} 
\\
$\rho_a = 0.3 \quad \rho_v = 0.6$ & $\rho_a = 0.5 \quad \rho_v = 0.5$ \\
\end{tabular}
\end{adjustbox}
\end{tabular}
}
\label{fig:appendix-v-pruning-ratio}
\vspace{-2.mm}

\hspace{0mm}

\vspace{1mm}
\captionsetup{font={small}}
    \caption{\textbf{More visualization of dynamic pruning ratios.} The figure illustrates how audio token retention guides the allocation of video token pruning. Specifically, for time windows with low audio retention, we allocate a higher video pruning ratio while maintaining a constant total pruning rate. }
    \label{fig:abs-rho-appendix}

\vspace{-4mm}
\end{figure*}

\begin{figure}[t]
\centering
\captionsetup{font={small}, skip=0pt}
\includegraphics[width=1\linewidth]{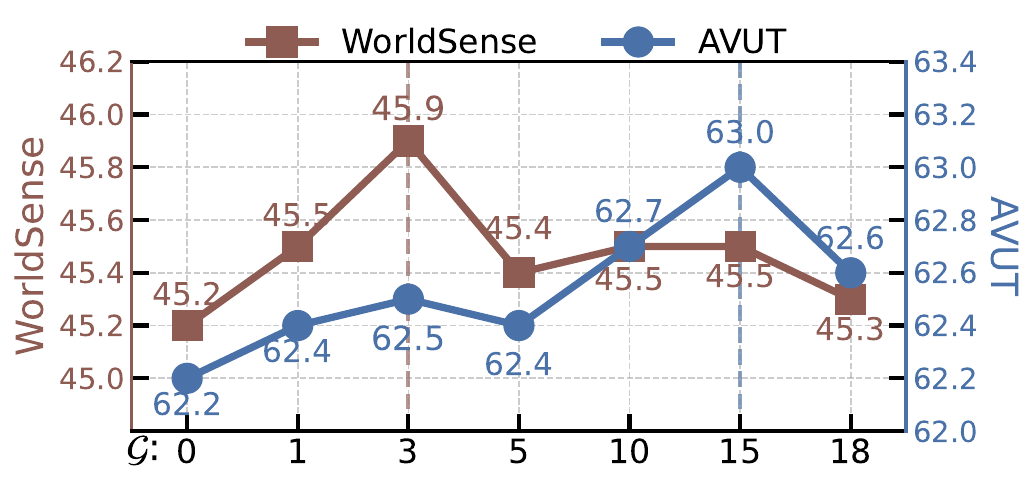}
\caption{\textbf{Ablation study on $\mathcal{G}$.} The accuracy of our method in a 45\% retained ratio is analyzed with the value of $\mathcal{G}$, which is defined as the number of tokens merged by each audio token anchor. All experiments illustrated in the figure were carried out on the Qwen2.5-Omni-7B model.}
\label{fig:analyses_g}
\vspace{-5mm}
\end{figure}

\section{More Experimental Results}
This section presents supplementary experimental results and ablation studies.

\cref{tab:WorldSense-appendix} presents comparison results under various pruning rates, primarily to further demonstrate that our method significantly outperforms other methods. 
Furthermore, OmniZip is designed to prune audio tokens more aggressively than video tokens (a heuristic derived from our analysis), but the data also demonstrates that our method's superior results are \emph{not solely dependent on this specific ratio}. 
For example, at a 50\% overall compression rate with a balanced 1:1 pruning ratio ($\rho_a$=0.5, $\rho_v$=0.5), OmniZip still achieves significantly better performance than other methods.

In addition, for the dynamic pruning ratio allocation, we provide more visualization results as shown in \cref{fig:abs-rho-appendix}.

\vspace{1mm}
\textbf{Ablation Study on $\mathcal{G}$.}
As shown in \cref{fig:analyses_g}, we evaluate the effect of $\mathcal{G}$. 
Primarily, the application of our audio token merging method yields substantial performance gains.
On the AVUT~\cite{yang2025audio}, which is \emph{audio-centric}, allocating a higher $\mathcal{G}$ proves to be appropriate. 
Conversely, in other benchmarks where audio is more balanced with video or serves as a supplementary modality, $\mathcal{G}=3$ achieves the best results, while larger values introduce noise and slightly degrade performance. 
This finding indicates that $\mathcal{G}$ can be dynamically tuned based on the task's reliance on audio information.

\section{Limitations and Future Work}
While this work is the first to demonstrate the acceleration of OmniLLMs via audio-visual token compression, it is important to acknowledge its current limitations. 
Firstly, the relative informational requirements of audio and video vary significantly across different tasks and contexts. 
Consequently, determining the optimal compression balance between audio and video tokens remains a significant challenge. 
Secondly, this method is designed primarily for offline inference and does not natively support online or arbitrary-length streaming audio-visual input \cite{qian2024streaming,di2025streaming,chen2025streamingtom,xu2025streamingvlm}. 
Developing a streaming video inference framework that effectively incorporates audio will be a primary focus of our future work.
Finally, the substantial parameter count of larger models continues to impede their practical deployment. Consequently, investigating how to combine token compression with other advanced efficiency techniques, such as model quantization \cite{lin2024awq,xiao2023smoothquant,liu2024spinquant,van2024gptvq} and pruning~\cite{frantar2023sparsegpt,xia2023sheared,sun2023simple,zhu2025obs}, represents a promising research direction.